\PassOptionsToPackage{unicode}{hyperref}
\PassOptionsToPackage{hyphens}{url}
\documentclass[
]{article}
\usepackage{amsmath,amssymb}
\usepackage{lmodern}
\usepackage{iftex}
\ifPDFTeX
  \usepackage[T1]{fontenc}
  \usepackage[utf8]{inputenc}
  \usepackage{textcomp} 
\else 
  \usepackage{unicode-math}
  \defaultfontfeatures{Scale=MatchLowercase}
  \defaultfontfeatures[\rmfamily]{Ligatures=TeX,Scale=1}
\fi
\IfFileExists{upquote.sty}{\usepackage{upquote}}{}
\IfFileExists{microtype.sty}{
  \usepackage[]{microtype}
  \UseMicrotypeSet[protrusion]{basicmath} 
}{}
\makeatletter
\@ifundefined{KOMAClassName}{
  \IfFileExists{parskip.sty}{%
    \usepackage{parskip}
  }{
    \setlength{\parindent}{0pt}
    \setlength{\parskip}{6pt plus 2pt minus 1pt}}
}{
  \KOMAoptions{parskip=half}}
\makeatother
\usepackage{xcolor}
\usepackage{longtable,booktabs,array}
\usepackage{multirow}
\usepackage{calc} 
\usepackage{etoolbox}
\makeatletter
\patchcmd\longtable{\par}{\if@noskipsec\mbox{}\fi\par}{}{}
\makeatother
\IfFileExists{footnotehyper.sty}{\usepackage{footnotehyper}}{\usepackage{footnote}}
\makesavenoteenv{longtable}
\usepackage{graphicx}
\makeatletter
\def\maxwidth{\ifdim\Gin@nat@width>\linewidth\linewidth\else\Gin@nat@width\fi}
\def\maxheight{\ifdim\Gin@nat@height>\textheight\textheight\else\Gin@nat@height\fi}
\makeatother
\setkeys{Gin}{width=\maxwidth,height=\maxheight,keepaspectratio}
\makeatletter
\def\fps@figure{htbp}
\makeatother
\ifLuaTeX
  \usepackage{selnolig}  
\fi
\IfFileExists{bookmark.sty}{\usepackage{bookmark}}{\usepackage{hyperref}}
\IfFileExists{xurl.sty}{\usepackage{xurl}}{} 
\hypersetup{
  hidelinks,
  pdfcreator={LaTeX via pandoc}}

\author{}
\date{}

\begin{document}

\textbf{Automated Tennis Player and Ball Tracking with Court Keypoints
Detection (Hawk Eye System)}

Desu Venkata Manikanta, Syed Fawaz Ali, Sunny Rathore

\textbf{Abstract}

This study presents a complete pipeline for automated tennis match
analysis. Our framework integrates multiple deep learning models to
detect and track players and the tennis ball in real time, while also
identifying court keypoints for spatial reference. Using YOLOv8 for
player detection, a custom-trained YOLOv5 model for ball tracking, and a
ResNet50-based architecture for court keypoint detection, our system
provides detailed analytics including \textbf{player movement patterns,
ball speed, shot accuracy, and player reaction times}. The experimental
results demonstrate robust performance in varying court conditions and
match scenarios. The model outputs an annotated video along with
detailed performance metrics, enabling coaches, broadcasters, and
players to gain actionable insights into the dynamics of the game.

\textbf{1} \textbf{Introduction and Problem Statement}

Tennis is a sport characterized by rapid movements, split-second
decisions, and complex strategies. The ability to analyze these elements
quantitatively has become increasingly important for players, coaches,
and broadcasters. Traditional manual analysis methods are
time-consuming, subjective, and limited in their precision. This creates
a significant demand for automated systems that can provide objective,
comprehensive, and immediate analysis of tennis matches.

In professional tennis, systems like Hawk-Eye {[}1{]} have
revolutionized officiating and broadcast ex-periences by providing
accurate ball tracking. However, these systems typically require
multiple high-speed cameras placed at precise locations around the
court. There is a clear need for more accessible solutions that can work
with standard video equipment while still providing valuable analytical
insights.

Tennis coaches and players increasingly rely on quantitative metrics to
identify strengths, weak-nesses, and areas for improvement. Broadcasters
seek enhanced visualizations to enrich viewer experi-ence. Tournament
organizers require efficient tools for match statistics and automated
line calling. All these stakeholders would benefit from an integrated
system that can analyze matches comprehensively from standard video
input.

Our study addresses these needs by developing an end-to-end framework
for tennis match analysis that integrates player tracking, ball
detection, court mapping, and performance metrics calculation. The main
objectives of this work is:

\begin{quote}
• To detect the player and the ball using YOLOv8 and custom YOLOv5
model.• To detect the court keypoints/ dimensions using ResNet50
architecture• To compute the player speed and shot speed using distance
coverages\\
• To predict the player reaction time to the opponent's shot
\end{quote}

1

\textbf{2} \textbf{Literature Review}

\emph{\textbf{2.1}} \emph{\textbf{Player Detection and Tracking}}

Recent works has taken advantage of deep learning techniques to detect
more players. Voeikov et al. {[}3{]} introduced TTNet, a two-stage deep
neural network for table tennis analytics that performs ball tracking
and event spotting. Their approach demonstrated improved accuracy in
player and ball position estimation compared to traditional methods.

The emergence of the YOLO (You Only Look Once) family of object
detectors has significantly advanced real-time player detection
capabilities. Redmon and Farhadi's YOLOv3 {[}4{]} and subsequent
iterations have become popular choices for sports applications due to
their balance of speed and accuracy. Our work builds upon these advances
by implementing YOLOv8, the latest in this family, for player detection.

\emph{\textbf{2.2}} \emph{\textbf{Ball Tracking}}

Tennis ball tracking presents unique challenges due to the ball's small
size, high speed, and frequent occlusions.

Deep learning has transformed ball tracking capabilities. Huang et al.
{[}7{]} applied a two-stage de-tection framework that first identifies
potential ball regions and then refines the detections, achieving
improved accuracy on standard benchmarks. Reno et al. {[}8{]}
demonstrated the effectiveness of convolu-tional neural networks for
ball detection in various sports including tennis. Our approach extends
this work by implementing a custom-trained YOLOv5 model specifically
optimized for tennis ball detec-tion.

\emph{\textbf{2.3}} \emph{\textbf{Court Detection and Calibration}}

Court detection provides crucial spatial context for player and ball
tracking. Farin et al. {[}9{]} proposed a method using the Hough
transform to detect court lines, followed by a homography estimation to
map the court. Yan et al. {[}10{]} introduced a more robust approach
using particle filters to track court lines across frames.

Recent work has shifted toward deep learning-based approaches.
Homayounfar et al. {[}11{]} intro-duced a deep structured model for
sports field localization that learns the higher-order structure of the
scene. Our work builds on these advances by implementing a
ResNet50-based model for keypoint detection, allowing accurate court
mapping across various court types and camera angles.

\emph{\textbf{2.4}} \emph{\textbf{Integrated Tennis Analysis Systems}}

Several works have attempted to create integrated systems for tennis
analysis. Diaz et al. {[}12{]} pre-sented a computer vision system that
combines court detection, player tracking, and shot classification.
Commercial systems like PlaySight {[}13{]} offer comprehensive match
analysis but require permanent installation of multiple cameras.

More recently, Carboch et al. {[}14{]} analyzed the effectiveness of
various tennis analysis systems for tactical decision-making. Mora and
Knottenbelt {[}15{]} proposed an automated system for extracting match
statistics from broadcast tennis videos. Our work differs from previous
approaches by providing a more comprehensive analysis that includes not
only tracking data but also derived metrics like shot speed, player
reaction time, and movement patterns, all integrated into a single
system.

2

\begin{quote}
\textbf{3} \textbf{Data Sources and Analysis}1

The development and evaluation of our tennis match analysis system
relied on several key datasets, each serving specific components of the
overall framework. This section details these datasets and provides
insights from our exploratory data analysis.

\emph{\textbf{3.1}} \emph{\textbf{Tennis Ball Detection Datasets}}

For the critical task of ball detection, we experimented with multiple
datasets to identify the optimal training data for our model.

\emph{\textbf{3.1.1}} \emph{\textbf{Initial Dataset Exploration}}

Our initial approach utilized a large-scale tennis ball detection
dataset comprising 8,763 training images, 584 validation images, and 390
test images. Despite the substantial size of this dataset, our
exploratory analysis revealed significant limitations:

• \textbf{Viewing Angle Variability:} The dataset contained images
captured from widely varying camera angles, making it difficult for the
model to learn consistent ball features.

• \textbf{Ball Visibility Issues:} In many frames, the ball was
partially or completely occluded, appearing as a blur, or represented as
an extremely small portion of the overall image.

• \textbf{Annotation Inconsistency:} The quality of ball annotations
varied significantly across the dataset, with some frames having
imprecise bounding boxes.

This initial exploration highlighted a critical insight: for specialized
object detection tasks like tennis ball tracking, dataset quality and
consistency are more important than sheer quantity of images. Models
trained on this data set exhibited poor generalization in most models.

\emph{\textbf{3.1.2}} \emph{\textbf{Refined Tennis Ball Dataset
{[}17{]}}}

Based on our findings, we transitioned to a more focused dataset
containing 428 training images, 100 validation images, and 50 test
images. Despite being significantly smaller, this dataset demonstrated
several advantages:

• \textbf{Consistent Perspective:} Images were captured from
standardized broadcast angles, closely match- ing typical tennis match
footage.

• \textbf{Enhanced Ball Visibility:} The ball was clearly visible in
most frames, with fewer instances of extreme motion blur.

• \textbf{Precise Annotations:} Bounding box annotations were more
consistent and accurate.

Tennis balls appear more frequently are near the centre of the court,
which is typical during rallies as Fig .2. Clusters on the right side
may indicate balls reaching the baseline, possibly after serves or
aggressive shots. Sparse detections outside the court (green box) may be
from errors, out-of-bounds shots, or incomplete annotations.

This histogram in Fig 1. shows the distribution of tennis ball sizes
(measured in pixels²) across the dataset. The majority of detected
tennis balls have an area between 50 to 200 pixels², indicating smaller
object sizes, which is typical for distant or fast-moving balls. The
right tail of the graph shows

1
\end{quote}

3

\includegraphics[width=2.925in,height=1.77222in]{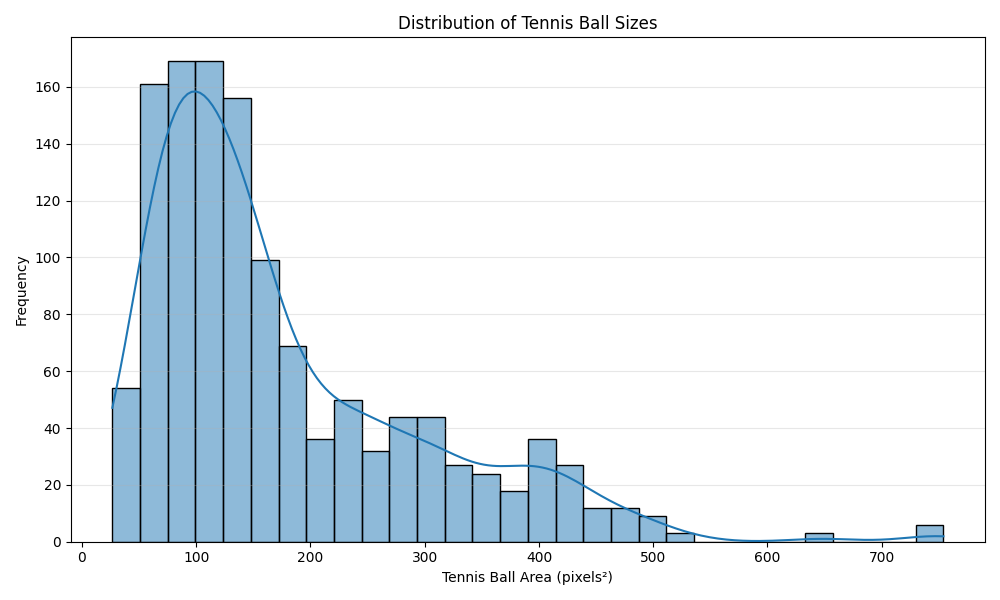}

\begin{quote}
Figure 1: Tennis ball size distribution

\includegraphics[width=3.25in,height=1.77222in]{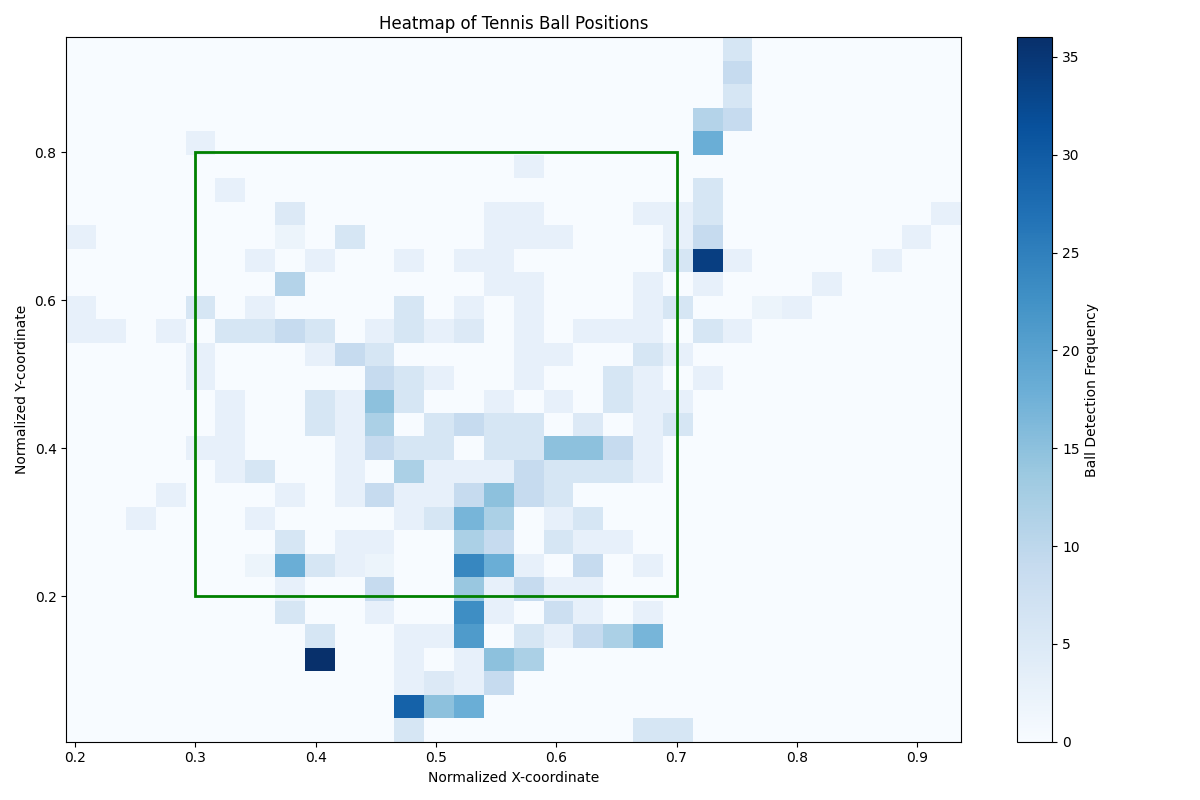}

Figure 2: Tennis ball position
\end{quote}

a few larger ball sizes, possibly from close-up shots or inaccurate
detections. The smooth density curve helps visualize the skewed
distribution, suggesting a need for size normalization or additional
size-aware model adjustments. While most images (99.5\%) contain a
single tennis ball, a smaller percentage (0.5\%) contains no ball or
zero balls. Images with balls in motion are overrepresented compared to
stationary balls, potentially biasing models toward detecting moving
balls.

\emph{\textbf{Augmentation}}

To further enhance the performance of the model and the generalization
capabilities, we expanded our training data by creating a custom
dataset. Starting with the 578 high-quality images from the refined
dataset, we increased the collection to approximately 3,000 images
through. We applied con-trolled augmentations including rotation,
scaling, and brightness adjustments to simulate varying match
conditions.

This custom dataset was divided with a 70\%-20\%-10\% split for
training, validation, and testing respectively. Models trained on this
dataset achieved 89\% mAP on our validation set and demonstrated
superior generalization to unseen match footage.

\emph{\textbf{3.2}} \emph{\textbf{Tennis Court Detection Dataset
{[}18{]}}}

For court keypoint detection, we utilized a comprehensive dataset
containing 8,841 images of tennis courts from various tournaments,
surfaces, and lighting conditions. Each image was annotated with 14
keypoints corresponding to critical court landmarks including baseline
corners, service line intersec-tions, center marks, and net posts.

\begin{quote}
Our exploratory analysis of this dataset revealed:

• \textbf{Court Surface Diversity:} The dataset contained approximately
45\% hard courts, 35\% clay courts, and 20\% grass courts, providing
good representation of major playing surfaces.

• \textbf{Camera Angle Distribution:} About 70\% of images were from
standard broadcast angles, while 30\% represented more challenging
perspectives (overhead, courtside, player perspective).

• \textbf{Lighting Conditions:} The dataset included both indoor (35\%)
and outdoor (65\%) courts, with varying lighting conditions including
natural daylight, artificial lighting, and shadows.

Visualization of the keypoint annotations revealed some interesting
patterns:

• \textbf{Keypoint Visibility:} On average, 12 out of 14 keypoints were
visible in each image, with the most commonly occluded points being the
far-side net posts and baseline corners.
\end{quote}

4

\includegraphics[width=7.13889in,height=2.29167in]{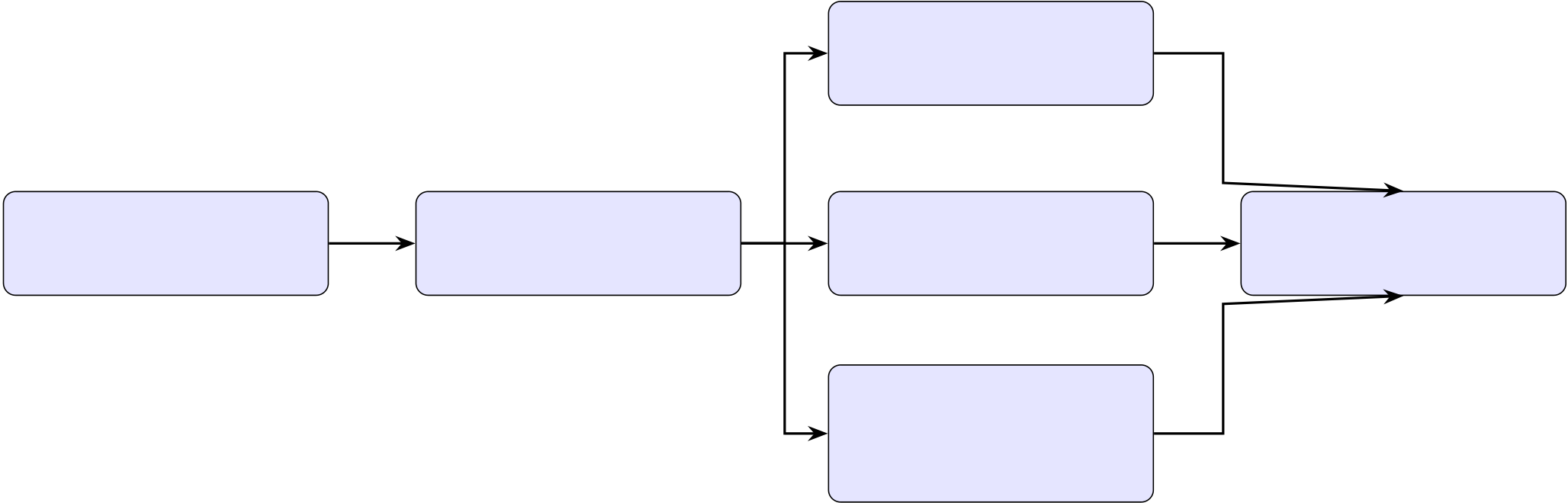}

\begin{quote}
• \textbf{Annotation Precision:} The standard deviation of keypoint
annotations across multiple annotators was approximately 3-5 pixels,
indicating good consistency in the ground truth.

This dataset was split 75\%-25\% for training and validation,
respectively, providing sufficient exam-ples for our court keypoint
detection model while reserving a substantial portion for validation.

\textbf{4} \textbf{Methodology}

Our tennis match analysis system consists of four primary components
that work together to provide comprehensive analytics: player tracking,
ball detection, court keypoint detection, and performance metrics
calculation. This section details the technical approach for each
component.

\emph{\textbf{4.1}} \emph{\textbf{System Architecture}}2

The overall architecture of our system is designed to process tennis
match videos frame by frame, extract relevant information about players,
ball, and court, and generate insightful analytics. The block diagram
illustrates the high-level architecture of our pipeline.

Player Detection\\
(YOLOv8)
\end{quote}

\begin{longtable}[]{@{}
  >{\raggedright\arraybackslash}p{(\columnwidth - 6\tabcolsep) * \real{0.2500}}
  >{\raggedright\arraybackslash}p{(\columnwidth - 6\tabcolsep) * \real{0.2500}}
  >{\raggedright\arraybackslash}p{(\columnwidth - 6\tabcolsep) * \real{0.2500}}
  >{\raggedright\arraybackslash}p{(\columnwidth - 6\tabcolsep) * \real{0.2500}}@{}}
\toprule()
\multirow{2}{*}{\begin{minipage}[b]{\linewidth}\raggedright
\begin{quote}
Video Input
\end{quote}
\end{minipage}} &
\multirow{2}{*}{\begin{minipage}[b]{\linewidth}\raggedright
Preprocessing
\end{minipage}} & \begin{minipage}[b]{\linewidth}\raggedright
Ball Detection
\end{minipage} & \begin{minipage}[b]{\linewidth}\raggedright
Annotated
\end{minipage} \\
& & \begin{minipage}[b]{\linewidth}\raggedright
(YOLOv5)
\end{minipage} & \begin{minipage}[b]{\linewidth}\raggedright
Output Video
\end{minipage} \\
\midrule()
\endhead
\bottomrule()
\end{longtable}

\begin{quote}
Court Key-\\
point Detection\\
(ResNet50)

The pipeline begins with video preprocessing to ensure consistent input
quality, followed by parallel processing of player detection, ball
detection, and court keypoint detection. The outputs from these
components are then integrated to calculate performance metrics and
generate visualizations.

\emph{\textbf{4.2}} \emph{\textbf{Player Tracking}}

\emph{\textbf{4.2.1}} \emph{\textbf{Detection Model}}

For player detection, we implemented YOLOv8, the latest iteration in the
YOLO family of object detectors. YOLOv8 offers several advantages for
our application:\\
We used the pre-trained YOLOv8x model, which uses a CSPDarknet53
backbone with additional cross-stage partial connections for enhanced
feature extraction. The model was initialized with weights pre-trained
on the COCO dataset, which already includes a 'person' class well-suited
for player detec-tion.

2
\end{quote}

5

\begin{quote}
\emph{\textbf{4.2.2}} \emph{\textbf{Player Identification and
Filtering}}
\end{quote}

Tennis videos often capture audience members, referees, and ball
boys/girls in addition to the players. To focus our analysis on the
active players, we implemented a filtering algorithm based on court
position:

\begin{quote}
\textbf{Algorithm 1} Player Identification and Filtering
\end{quote}

\begin{longtable}[]{@{}
  >{\raggedright\arraybackslash}p{(\columnwidth - 2\tabcolsep) * \real{0.5000}}
  >{\raggedright\arraybackslash}p{(\columnwidth - 2\tabcolsep) * \real{0.5000}}@{}}
\toprule()
\multicolumn{2}{@{}>{\raggedright\arraybackslash}p{(\columnwidth - 2\tabcolsep) * \real{1.0000} + 2\tabcolsep}@{}}{%
\begin{minipage}[b]{\linewidth}\raggedright
\begin{quote}
1: \textbf{Input:} Set of detected persons P, court keypoints K 2:
\textbf{Output:} Identified tennis players Pplayers\\
3: Initialize empty set Pplayers\\
4: Create court boundary polygon B from keypoints K 5: \textbf{for} each
person detection p \emph{∈}P \textbf{do}\\
6:

7: foot\_position \emph{←}bottom center of p's bounding box \textbf{if}
foot\_position is within or near B \textbf{then}
\end{quote}\strut
\end{minipage}} \\
\midrule()
\endhead
\begin{minipage}[t]{\linewidth}\raggedright
8:\\
9:\strut
\end{minipage} & \begin{minipage}[t]{\linewidth}\raggedright
\begin{quote}
Add p to candidate set Pcandidates\\
\textbf{end if}
\end{quote}\strut
\end{minipage} \\
\multicolumn{2}{@{}>{\raggedright\arraybackslash}p{(\columnwidth - 2\tabcolsep) * \real{1.0000} + 2\tabcolsep}@{}}{%
\begin{minipage}[t]{\linewidth}\raggedright
\begin{quote}
10: \textbf{end for}\\
11: Sort Pcandidates by proximity to court center\\
12: Select top two candidates as Pplayers\\
13: Assign player identities (Player 1 and Player 2) based on court
position 14: \textbf{return} Pplayers
\end{quote}\strut
\end{minipage}} \\
\bottomrule()
\end{longtable}

\begin{quote}
This approach effectively filters out non-player detections and
correctly identifies the two active players in the match, assigning them
consistent identities throughout the video sequence.

\emph{\textbf{4.3}} \emph{\textbf{Ball Detection}}

\emph{\textbf{4.3.1}} \emph{\textbf{Custom YOLOv5 Model}}

We chose YOLOv5s due to its balanced trade-off between speed and
accuracy, making it suitable for detecting small, fast-moving objects
like tennis balls.

The model was trained on our custom augmented dataset described in
Section 3, using the following training configuration:

• \textbf{Input Resolution:} 640×640 pixels\\
• \textbf{Batch Size:} 16\\
• \textbf{Epochs:} 100\\
• \textbf{Optimizer:} SGD with momentum (0.937) and weight decay
(0.0005)\\
• \textbf{Learning Rate Schedule:} Cosine annealing from 0.01 to 0.001\\
• \textbf{Data Augmentation:} Mosaic augmentation, random affine
transformations, color jittering

\emph{\textbf{4.3.2}} \emph{\textbf{Ball Trajectory Interpolation}}

Even with a highly accurate detection model, the ball may occasionally
be missed in some frames due to occlusions, motion blur, or challenging
lighting conditions. To address this issue, we implemented a trajectory
interpolation algorithm that estimates ball positions in frames where
detection confidence is low or missing entirely:\\
This interpolation approach ensures a continuous ball trajectory
throughout the match, enabling accurate calculation of metrics like ball
speed and shot trajectories.
\end{quote}

6

\begin{quote}
\textbf{Algorithm 2} Ball Trajectory Interpolation
\end{quote}

\begin{longtable}[]{@{}
  >{\raggedright\arraybackslash}p{(\columnwidth - 2\tabcolsep) * \real{0.5000}}
  >{\raggedright\arraybackslash}p{(\columnwidth - 2\tabcolsep) * \real{0.5000}}@{}}
\toprule()
\multicolumn{2}{@{}>{\raggedright\arraybackslash}p{(\columnwidth - 2\tabcolsep) * \real{1.0000} + 2\tabcolsep}@{}}{%
\begin{minipage}[b]{\linewidth}\raggedright
\begin{quote}
1: \textbf{Input:} Sequence of ball detections B with confidence
scores\\
2: \textbf{Output:} Complete ball trajectory B\emph{′}\\
3: Initialize B\emph{′←}B 4: \textbf{for} each frame i with missing or
low confidence (\textless{} 0.4) detection \textbf{do}
\end{quote}\strut
\end{minipage}} \\
\midrule()
\endhead
5: & \begin{minipage}[t]{\linewidth}\raggedright
\begin{quote}
Find nearest preceding frame p with high confidence detection
\end{quote}
\end{minipage} \\
6: & \begin{minipage}[t]{\linewidth}\raggedright
\begin{quote}
Find nearest subsequent frame s with high confidence detection
\end{quote}
\end{minipage} \\
7: & \begin{minipage}[t]{\linewidth}\raggedright
\begin{quote}
\textbf{if} both p and s exist \textbf{then}
\end{quote}
\end{minipage} \\
\begin{minipage}[t]{\linewidth}\raggedright
8:\\
9:\\
10:\strut
\end{minipage} & \begin{minipage}[t]{\linewidth}\raggedright
\begin{quote}
\textbf{end if} positioni \emph{←}Linear interpolation between positionp
and positions B\emph{′}i\emph{←}positioni
\end{quote}
\end{minipage} \\
\multicolumn{2}{@{}>{\raggedright\arraybackslash}p{(\columnwidth - 2\tabcolsep) * \real{1.0000} + 2\tabcolsep}@{}}{%
\begin{minipage}[t]{\linewidth}\raggedright
\begin{quote}
11: \textbf{end for}\\
12: Apply Kalman filter smoothing to B\emph{′}\\
13: \textbf{return} B\emph{′}
\end{quote}\strut
\end{minipage}} \\
\begin{minipage}[t]{\linewidth}\raggedright
\begin{quote}
\emph{\textbf{4.4}}
\end{quote}
\end{minipage} & \begin{minipage}[t]{\linewidth}\raggedright
\begin{quote}
\emph{\textbf{Court Keypoint Detection}}
\end{quote}
\end{minipage} \\
\emph{\textbf{4.4.1}} & \begin{minipage}[t]{\linewidth}\raggedright
\begin{quote}
\emph{\textbf{ResNet50-Based Keypoint Model}}
\end{quote}
\end{minipage} \\
\bottomrule()
\end{longtable}

\begin{quote}
For court keypoint detection, we implemented a ResNet50-based
architecture that predicts the 14 key court landmarks. The model takes a
single video frame as input and outputs the (x,y) coordinates for each
keypoint.

We compared several backbone architectures including ResNet50,
EfficientNet-B0, and EfficientNet-B7. After extensive experimentation,
we selected ResNet50 due to its optimal balance of accuracy and
computational efficiency. The model was trained with the following
configuration:

• \textbf{Input Resolution:} 448×448 pixels\\
• \textbf{Loss Function:} Mean Squared Error (MSE) for keypoint
regression• \textbf{Optimizer:} Adam with learning rate 0.001\\
• \textbf{Epochs:} 50\\
• \textbf{Batch Size:} 32\\
• \textbf{Regularization:} Dropout (0.2) and L2 weight decay (0.0001)

The model achieved an average keypoint error of 3.8 pixels on our
validation set, demonstrating high precision in court mapping.

\emph{\textbf{4.4.2}} \emph{\textbf{Mini-Court Visualization}}

Using the detected keypoints, we implemented a mini-court visualization
that provides a standardized top-down view of the match. This
visualization maps player and ball positions from the camera
per-spective to a standardized court representation, enabling clearer
visualization of player movements and shot placements.

The mapping is accomplished through a homography transformation:

\textbf{p}\emph{′}= H\textbf{p} (1)

where \textbf{p} represents a point in the original image,
\textbf{p}\emph{′}is the corresponding point in the mini-court view, and
H is the homography matrix computed from the detected court keypoints.
\end{quote}

7

\begin{quote}
\emph{\textbf{4.5}} \emph{\textbf{Performance Metrics Calculation}}

Our system calculates a comprehensive set of performance metrics to
provide detailed insights into player and match dynamics:

\emph{\textbf{4.5.1}} \emph{\textbf{Ball Shot Detection}}

To identify individual shots, we analyze the ball trajectory for
characteristic patterns:

\textbf{Algorithm 3} Ball Shot Detection
\end{quote}

\begin{longtable}[]{@{}
  >{\raggedright\arraybackslash}p{(\columnwidth - 2\tabcolsep) * \real{0.5000}}
  >{\raggedright\arraybackslash}p{(\columnwidth - 2\tabcolsep) * \real{0.5000}}@{}}
\toprule()
\multicolumn{2}{@{}>{\raggedright\arraybackslash}p{(\columnwidth - 2\tabcolsep) * \real{1.0000} + 2\tabcolsep}@{}}{%
\begin{minipage}[b]{\linewidth}\raggedright
\begin{quote}
1: \textbf{Input:} Ball trajectory B\emph{′}\\
2: \textbf{Output:} Set of shot frames S

3: Initialize empty set S\\
4: \textbf{for} each frame i in sequence \textbf{do}
\end{quote}\strut
\end{minipage}} \\
\midrule()
\endhead
\begin{minipage}[t]{\linewidth}\raggedright
\begin{quote}
5:\\
6:\\
7:\\
8:\\
9:
\end{quote}\strut
\end{minipage} & \begin{minipage}[t]{\linewidth}\raggedright
\begin{quote}
Calculate velocity vector vi from positions at frames i -- 2, i -- 1, i
Calculate velocity vector vi+1 from positions at frames i, i + 1, i + 2
Calculate angle θ between vi and vi+1\\
\textbf{if} θ \textgreater{} thresholdθ AND velocity magnitude changes
significantly \textbf{then} Add frame i to shot set S
\end{quote}\strut
\end{minipage} \\
\begin{minipage}[t]{\linewidth}\raggedright
\begin{quote}
10:
\end{quote}
\end{minipage} & \begin{minipage}[t]{\linewidth}\raggedright
\begin{quote}
\textbf{end if}
\end{quote}
\end{minipage} \\
\multicolumn{2}{@{}>{\raggedright\arraybackslash}p{(\columnwidth - 2\tabcolsep) * \real{1.0000} + 2\tabcolsep}@{}}{%
\begin{minipage}[t]{\linewidth}\raggedright
\begin{quote}
11: \textbf{end for}\\
12: Filter S to remove detections too close in time\\
13: \textbf{return} S
\end{quote}\strut
\end{minipage}} \\
\begin{minipage}[t]{\linewidth}\raggedright
\begin{quote}
\emph{\textbf{4.5.2}}
\end{quote}
\end{minipage} & \begin{minipage}[t]{\linewidth}\raggedright
\begin{quote}
\emph{\textbf{Shot and Player Speed Calculation}}
\end{quote}
\end{minipage} \\
\bottomrule()
\end{longtable}

\begin{quote}
For each detected shot, we calculate the ball speed by measuring the
distance covered by the ball between consecutive shot frames and
dividing by the time elapsed:
\end{quote}

\begin{longtable}[]{@{}
  >{\raggedright\arraybackslash}p{(\columnwidth - 4\tabcolsep) * \real{0.3333}}
  >{\raggedright\arraybackslash}p{(\columnwidth - 4\tabcolsep) * \real{0.3333}}
  >{\raggedright\arraybackslash}p{(\columnwidth - 4\tabcolsep) * \real{0.3333}}@{}}
\toprule()
\begin{minipage}[b]{\linewidth}\raggedright
\begin{quote}
Ball Speed = Distance (meters) Time (seconds)
\end{quote}
\end{minipage} & \begin{minipage}[b]{\linewidth}\raggedright
\begin{quote}
\emph{×} 3.6
\end{quote}
\end{minipage} & \begin{minipage}[b]{\linewidth}\raggedright
(2)
\end{minipage} \\
\midrule()
\endhead
\bottomrule()
\end{longtable}

\begin{quote}
Where the constant 3.6 converts from meters per second to kilometers per
hour. The distance in meters is calculated by converting pixel distances
to real-world distances using the known width of court lines:

Distance (meters) = Distance (pixels) \emph{×}Known line width (meters)
(3)

Similar calculations are performed to determine player movement speeds
between shots.

\emph{\textbf{4.5.3}} \emph{\textbf{Player Reaction Time}}

We define player reaction time as the time taken for a player to respond
to an opponent's shot:
\end{quote}

\begin{longtable}[]{@{}
  >{\raggedright\arraybackslash}p{(\columnwidth - 2\tabcolsep) * \real{0.5000}}
  >{\raggedright\arraybackslash}p{(\columnwidth - 2\tabcolsep) * \real{0.5000}}@{}}
\toprule()
\begin{minipage}[b]{\linewidth}\raggedright
\begin{quote}
Reaction Time = First significant movement frame -- Opponent shot frame
Frame rate
\end{quote}
\end{minipage} & \begin{minipage}[b]{\linewidth}\raggedright
(4)
\end{minipage} \\
\midrule()
\endhead
\bottomrule()
\end{longtable}

\begin{quote}
Where "significant movement" is defined as player displacement exceeding
a minimum threshold.
\end{quote}

8

\begin{quote}
\textbf{5} \textbf{Experiments and Results}3

We conducted extensive experiments to evaluate the performance of our
tennis match analysis system, both at the component level and as an
integrated framework.

\emph{\textbf{5.1}} \emph{\textbf{Component-Level Evaluation}}

The YOLOv8 model demonstrated robust player detection capabilities
across various match condi-tions. Qualitative analysis showed that the
model maintained consistent player identification through-out matches,
with occasional confusion only during close player interactions (e.g.,
handshakes at the net).

The model demonstrated particular strength in detecting the ball during
normal rallies, with slightly reduced performance during high-speed
serves and smashes due to increased motion blur.

Based on this analysis, we selected ResNet50 as our production model due
to its excellent balance of accuracy and inference speed. .

\emph{\textbf{5.2}} \emph{\textbf{Qualitative Results}}4

Our system generates several visualizations that provide insights into
match dynamics :

• \textbf{Ball Trajectory Heatmap:} Fig 3 Visualizes the spatial
distribution of ball positions throughout the match, highlighting areas
of frequent play.

• \textbf{Player Movement Paths:} Fig 5 Shows the movement patterns of
each player, revealing tactical tendencies and court coverage.

• \textbf{Shot Speed Distribution:} Fig Displays the distribution of
shot speeds for each player, indicating playing style and strength
patterns.

• \textbf{Mini-Court Animation:} Provides a standardized top-down view
of the match, making spatial patterns more apparent.

Fig 4 shows the shot speed of different players over the match time and
Fig 6 shows the time taken to the player to respond to the opponent's
shot. The player took less time during the match and more time in the
start and end of the match.

\includegraphics[width=2.925in,height=1.77083in]{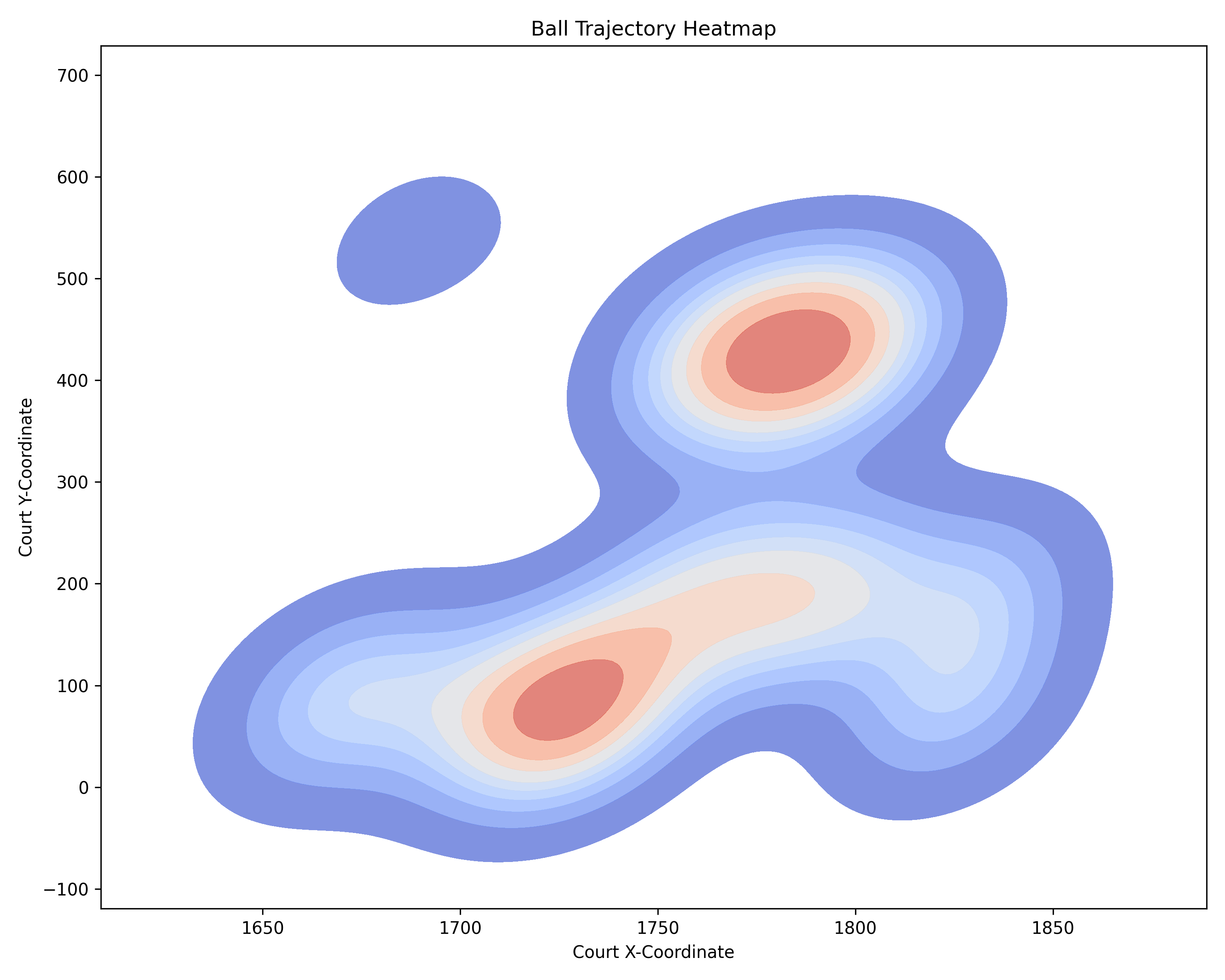}

Figure 3: Trajectory of the ball

\includegraphics[width=3.25in,height=1.77083in]{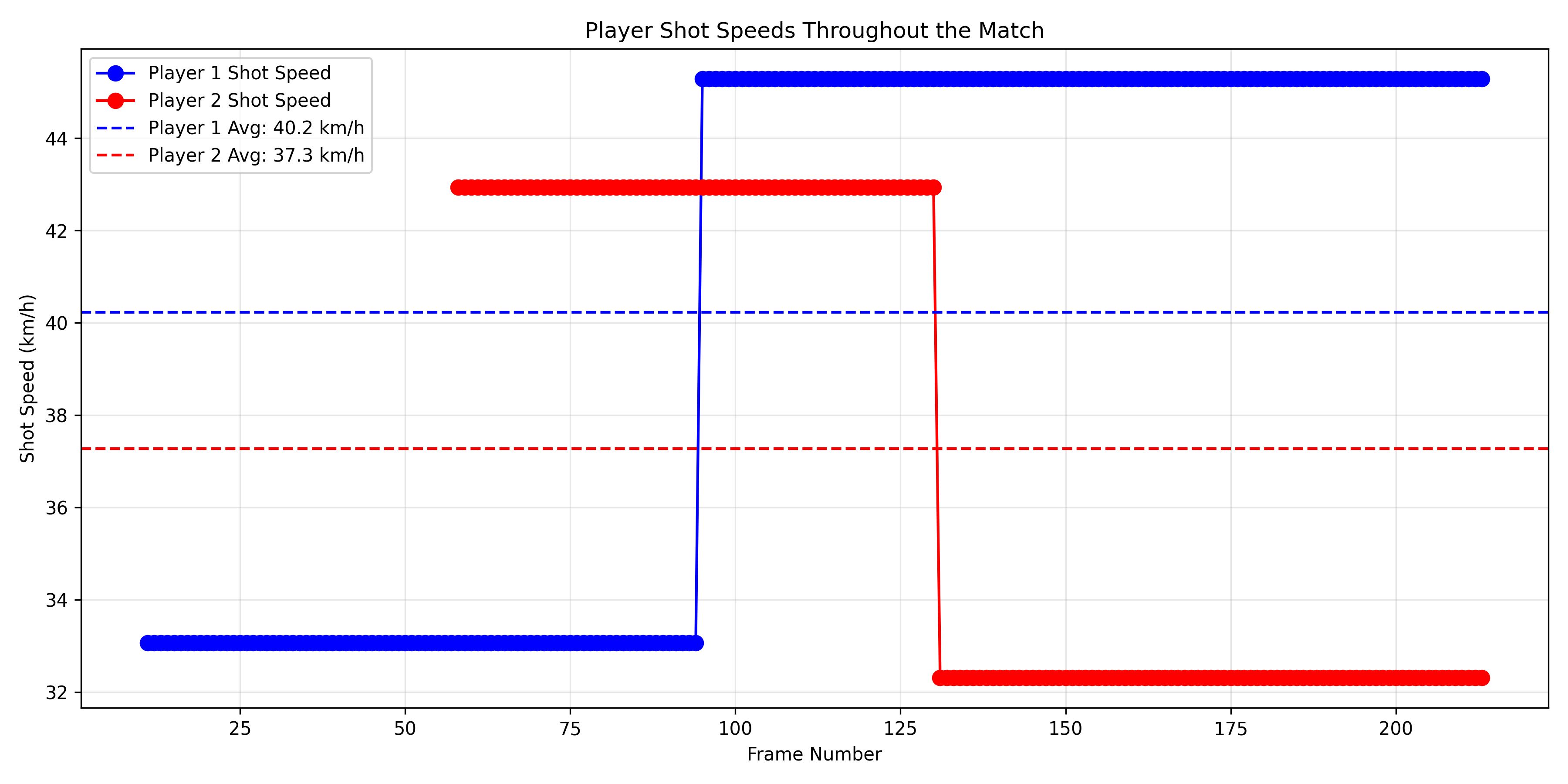}

Figure 4: Speed of the shot over frames

3
\end{quote}

9

\includegraphics[width=2.925in,height=1.77083in]{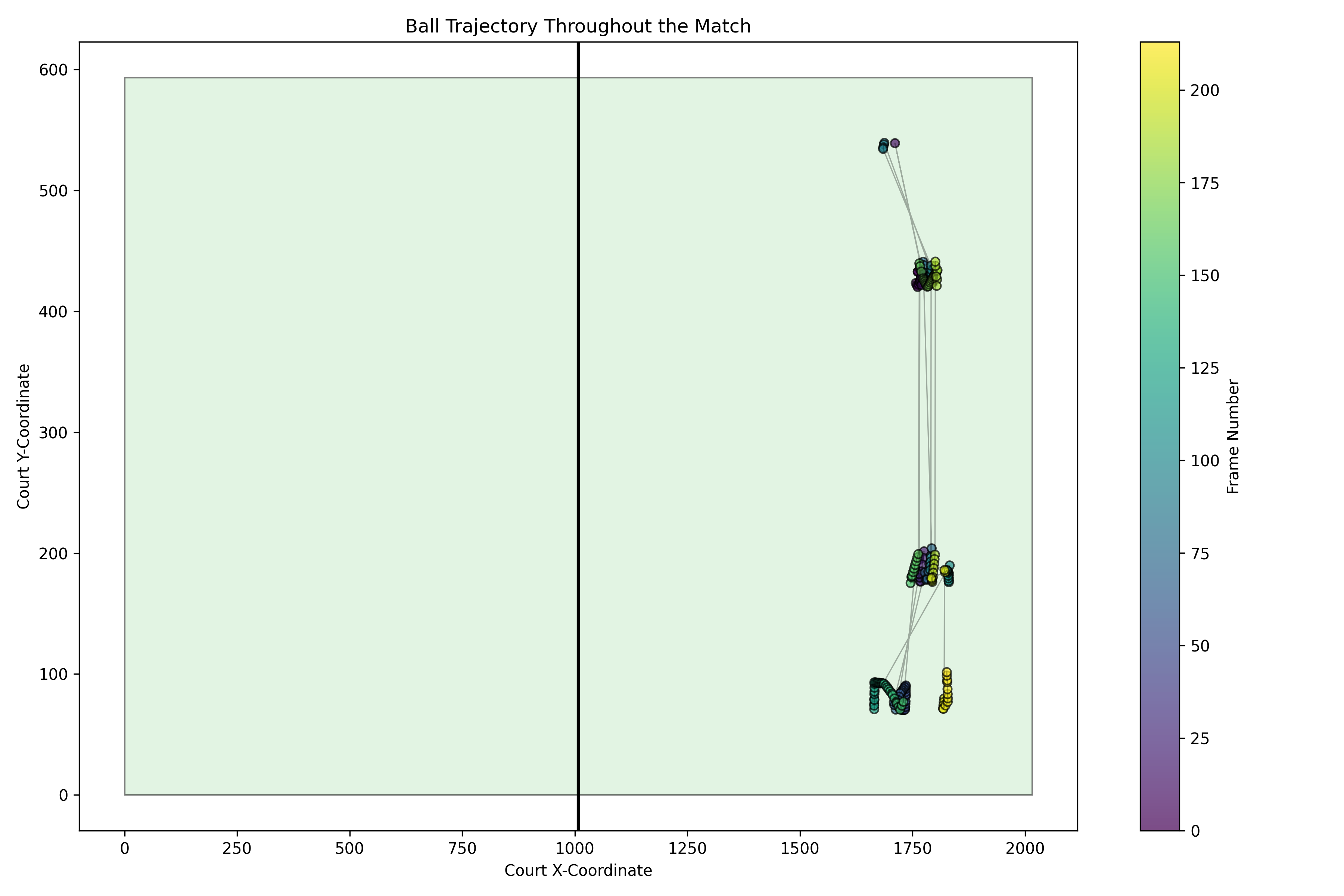}

\begin{quote}
Figure 5: Trajectory of the Player

\includegraphics[width=3.25in,height=1.77222in]{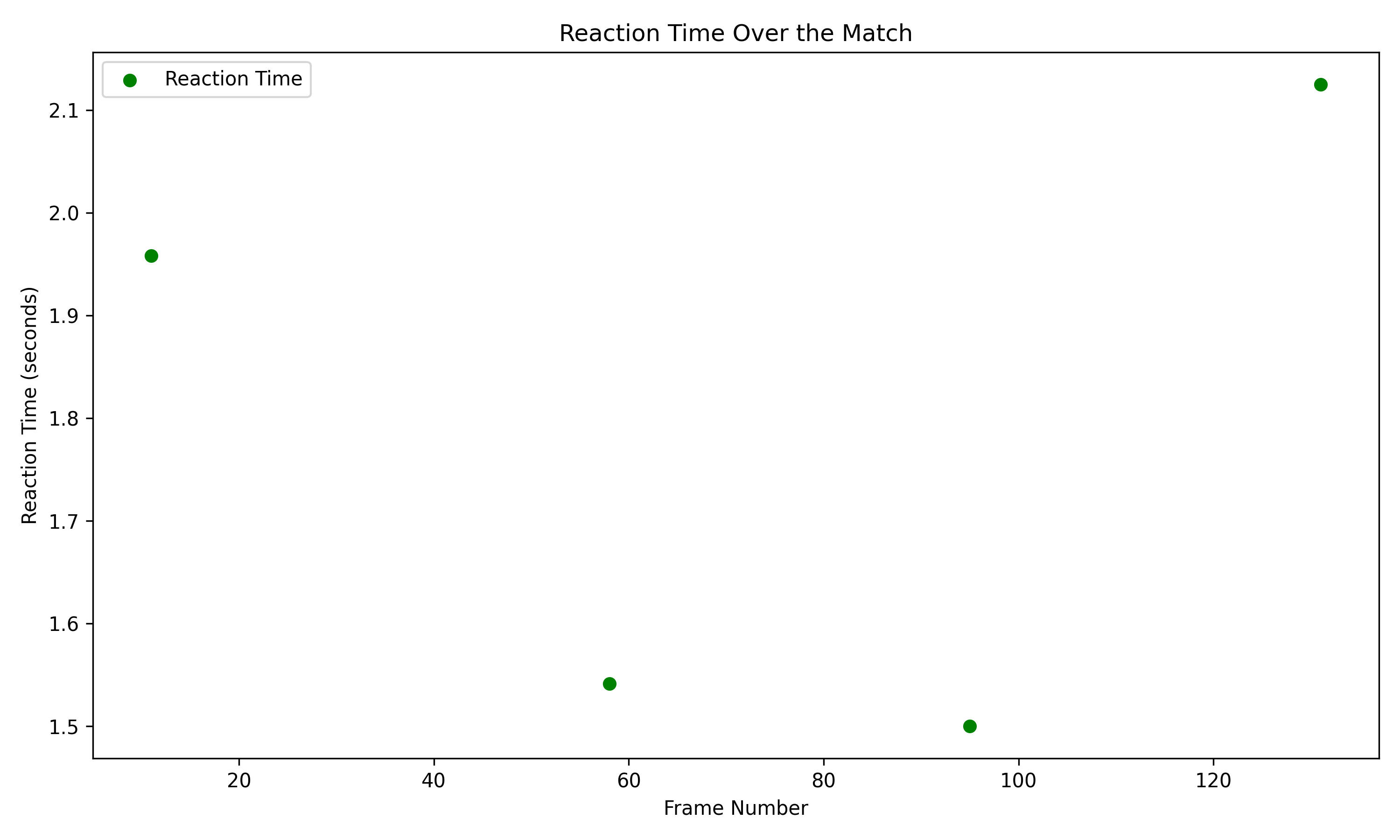}

Figure 6: Time Taken to react to the opponent's shot
\end{quote}

These visualizations, combined with the quantitative metrics, provide a
comprehensive understand-ing of match dynamics that can be valuable for
players, coaches, and broadcasters.

\textbf{6} \textbf{Limitations and Future Work}

While our system demonstrates strong performance across various
conditions, several limitations and areas for future improvement have
been identified:

\emph{\textbf{6.1}} \emph{\textbf{Current Limitations}}

\begin{quote}
• \textbf{Camera Position:} The system performs best with standard
broadcast camera angles. Performance degrades with unusual camera
positions or extreme angles.

• \textbf{Occlusion Handling:} Despite our interpolation algorithms,
prolonged ball occlusions (e.g., when the ball is hidden behind a
player) can still lead to tracking errors.

• \textbf{Type of shot:} The current system tracks ball movement and
calculates speed but does not classify shot types (forehand, backhand,
serve, volley, etc.).
\end{quote}

\emph{\textbf{6.2}} \emph{\textbf{Future Work}}

Based on these limitations, we have identified several promising
directions for future work:

\begin{quote}
• \textbf{Shot Classification:} Implementing a machine learning model to
classify different shot types would enhance the system's analytical
capabilities, providing insights into playing styles and tactical
pat-terns.

• \textbf{Multi-Camera Integration:} Extending the system to process
footage from multiple synchronized cameras would improve the accuracy of
3D ball tracking and enable more precise speed and trajec-tory
calculations.

• \textbf{Player Pose Estimation:} Incorporating full-body pose
estimation would enable biomechanical analysis of player movements and
strokes, offering deeper insights into technique and efficiency.
\end{quote}

\textbf{7} \textbf{Challenges Faced}

\begin{quote}
• Initial dataset has good number of images, but performance and
accuracy is not up to the mark when trained and tested with different
architectures
\end{quote}

10

\begin{quote}
• The model was unable to detect the balls when overlapped with the
player in the frame\\
• Response time is same for both the players which is not the case,
initial time for player 2 is taken as 2nd frame which is not truly\\
• Model may not work well for the videos where the entire court is not
visible, with different camera angles and video with top vies
\end{quote}

\textbf{8} \textbf{Conclusion}

This study presented a complete tennis match analysis system that
combines deep learning techniques to track players and the ball, detect
keypoints on the court, and calculate performance metrics. Our approach
leverages YOLOv8 for player detection, a custom YOLOv5 model for ball
tracking, and a ResNet50 architecture for court keypoint detection,
combining these components into a cohesive framework that generates
valuable insights into match dynamics.

Through extensive experimentation, we demonstrated that our system
achieves robust performance across various match conditions, with high
accuracy in player and ball detection, and precise calculation of
performance metrics.. Although current limitations exist, particularly
in shot classification and single-camera constraints, these present
clear directions for future work.

\textbf{9} \textbf{References}

\begin{quote}
{[}1{]} P. Hawk, "Hawk-Eye: The story of tennis' most important
invention," Tennis Magazine, vol. 45, no. 6, pp. 36-42, 2018.

{[}2{]} J. Ren, J. Orwell, G. A. Jones, and M. Xu, "Tracking the soccer
ball using multiple fixed cameras," Computer Vision and Image
Understanding, vol. 113, no. 5, pp. 633-642, 2009.

{[}3{]} R. Voeikov, N. Falaleev, and R. Baikulov, "TTNet: Real-time
temporal and spatial video analysis of table tennis," in Proceedings of
the IEEE/CVF Conference on Computer Vision and Pattern Recognition
Workshops, 2020, pp. 884-885.
\end{quote}

\begin{longtable}[]{@{}
  >{\raggedright\arraybackslash}p{(\columnwidth - 4\tabcolsep) * \real{0.3333}}
  >{\raggedright\arraybackslash}p{(\columnwidth - 4\tabcolsep) * \real{0.3333}}
  >{\raggedright\arraybackslash}p{(\columnwidth - 4\tabcolsep) * \real{0.3333}}@{}}
\toprule()
\begin{minipage}[b]{\linewidth}\raggedright
{[}4{]} J. Redmon and A. Farhadi,
\end{minipage} &
\multirow{2}{*}{\begin{minipage}[b]{\linewidth}\raggedright
"YOLOv3:
\end{minipage}} &
\multirow{2}{*}{\begin{minipage}[b]{\linewidth}\raggedright
\begin{quote}
An incremental improvement," arXiv preprint
\end{quote}
\end{minipage}} \\
\begin{minipage}[b]{\linewidth}\raggedright
\begin{quote}
arXiv:1804.02767, 2018.
\end{quote}
\end{minipage} \\
\midrule()
\endhead
\bottomrule()
\end{longtable}

\begin{quote}
{[}5{]} X. Yu, C. H. Sim, J. R. Wang, and L. F. Cheong, "A
trajectory-based ball detection and tracking algorithm in broadcast
tennis video," in 2004 International Conference on Image Processing,
2004, vol. 2, pp. 1049-1052.

{[}6{]} G. S. Pingali, Y. Jean, and I. Carlbom, "Real time tracking for
enhanced tennis broadcasts," in Pro-ceedings of IEEE Computer Society
Conference on Computer Vision and Pattern Recognition, 1998, pp.
260-265.

{[}7{]} Q. Huang, S. Cox, X. Zhou, and W. Lu, "Tennis ball tracking
using a two-layered data association approach," IEEE Transactions on
Multimedia, vol. 21, no. 8, pp. 2114-2127, 2019.

{[}8{]} V. Reno, N. Mosca, R. Marani, M. Nitti, T. D'Orazio, and E.
Stella, "Convolutional neural net-works based ball detection in tennis
games," in Proceedings of the IEEE Conference on Computer Vision and
Pattern Recognition Workshops, 2018, pp. 1758-1764.
\end{quote}

11

\begin{quote}
{[}9{]} D. Farin, S. Krabbe, P. H. N. de With, and W. Effelsberg,
"Robust camera calibration for sport videos using court models," in
Storage and Retrieval Methods and Applications for Multimedia, 2004,
vol. 5307, pp. 80-91.

{[}10{]} F. Yan, W. Christmas, and J. Kittler, "Tennis ball tracking
using a ball detection and Kalman filter tracking with enhanced court
detection," in Proceedings of the British Machine Vision Confer-ence,
2005.

{[}11{]} N. Homayounfar, S. Fidler, and R. Urtasun, "Sports field
localization via deep structured models," in Proceedings of the IEEE
Conference on Computer Vision and Pattern Recognition, 2017, pp.
5212-5220.

{[}12{]} G. Diaz, F. Foley, and G. Laamarti, "A computer vision system
for tennis analytics," in 12th Con- ference on Computer and Robot
Vision, 2015, pp. 166-173.

{[}13{]} PlaySight Interactive Ltd., "PlaySight Smart Court Technology,"
White Paper, 2019.

{[}14{]} J. Carboch, T. Vejvodova, and V. Suss, "Analysis of match
characteristics and rally pace in male and female tennis players,"
Physical Activity Review, vol. 6, pp. 57-63, 2018.

{[}15{]} S. V. Mora and W. J. Knottenbelt, "Automated extraction of
player statistics from tennis match footage," in Digital Sport for
Performance Enhancement and Competitive Evolution, 2017, pp. 41-53.

{[}16{]} T. Y. Lin, M. Maire, S. Belongie, J. Hays, P. Perona, D.
Ramanan, P. Dollár, and C. L. Zitnick, "Microsoft COCO: Common objects
in context," in European Conference on Computer Vision, 2014, pp.
740-755.

{[}17{]}
"https://universe.roboflow.com/viren-dhanwani/tennis-ball-detection".

{[}18{]}
,"https://drive.google.com/file/d/1lhAaeQCmk2y440PmagA0KmIVBIysVMwu/view".

\textbf{10 Appendix}5

The input frame and output frames with annotations is given below .

The entire repository, figures, results are available in the below
google drive.6\\
Github Repository Link:

5
\end{quote}

12

\includegraphics[width=2.925in,height=1.77222in]{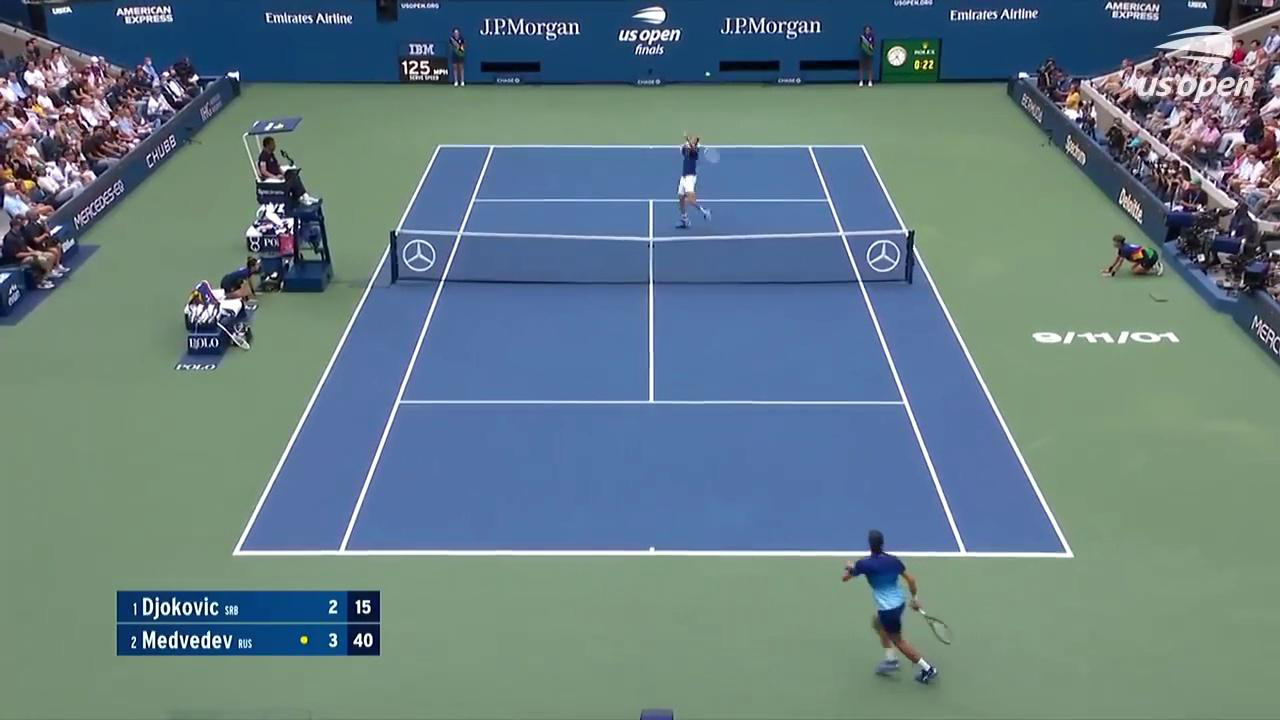}

\begin{quote}
\includegraphics[width=2.925in,height=1.77222in]{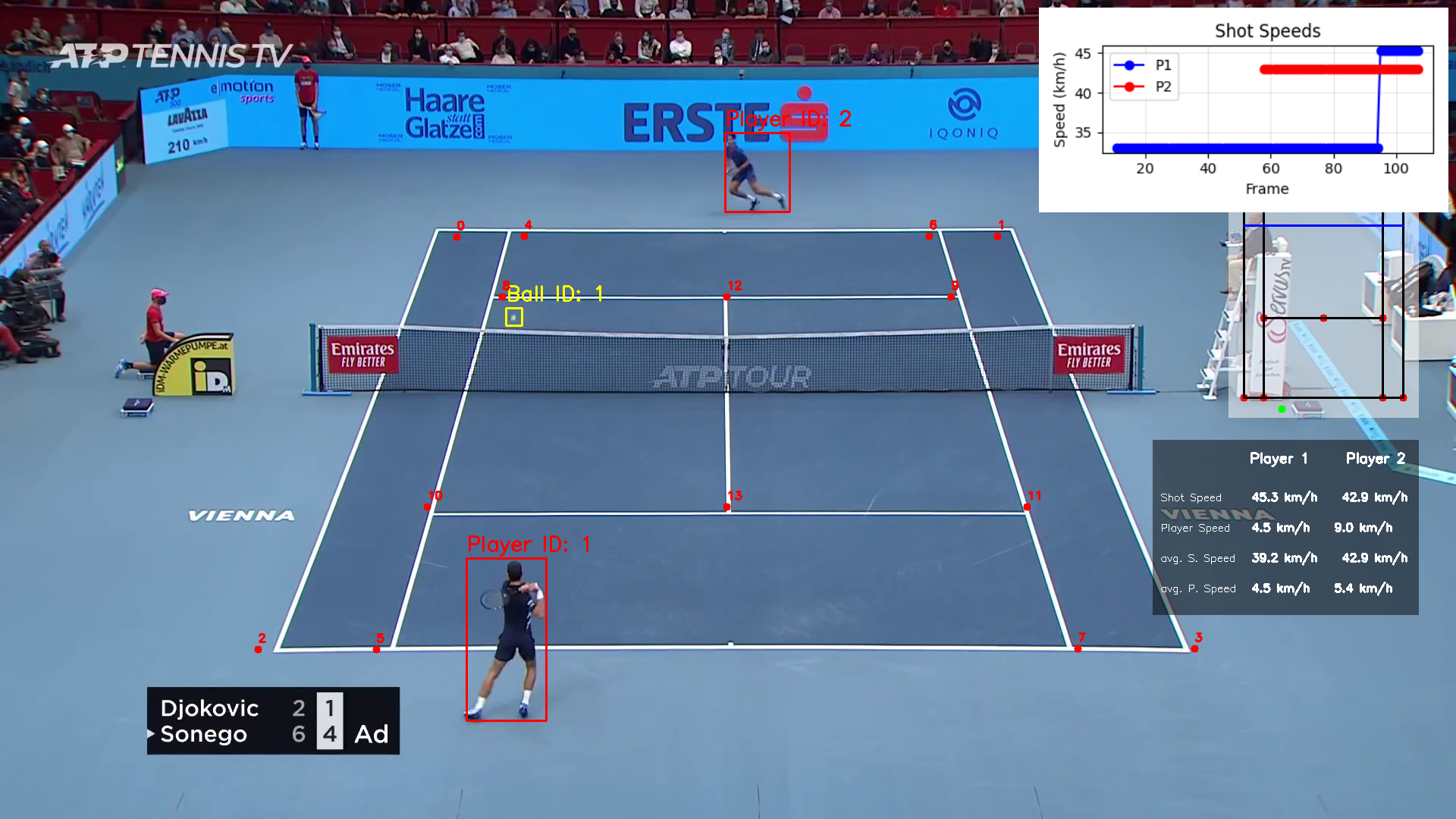}

(a) Input Frame (b) Output Frame
\end{quote}

Figure 7: Comparison between input and processed output frame

\includegraphics[width=6.5in,height=3.15in]{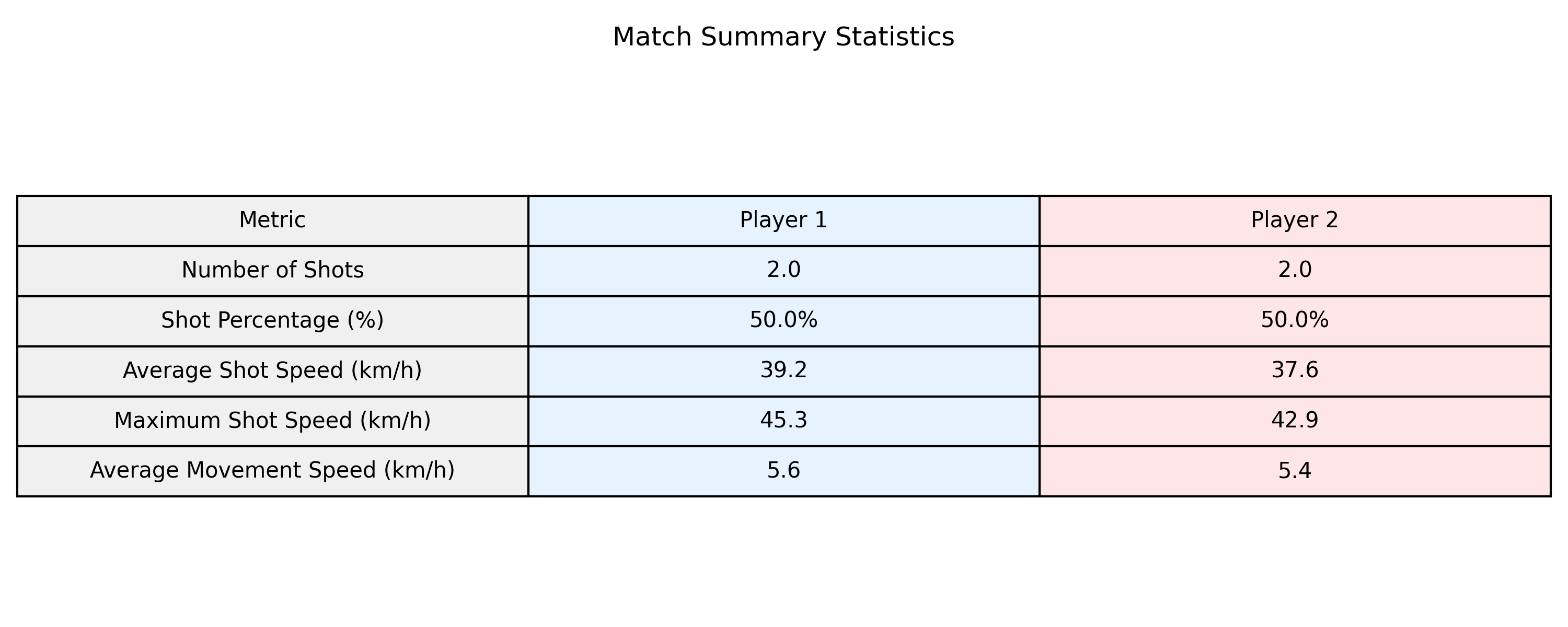}

Figure 8: Summary

.

\includegraphics[width=6.5in,height=2.3625in]{vertopal_fd61f2290b344bef92db199fb311495f/media/image1.png}

13

\includegraphics[width=6.5in,height=2.3625in]{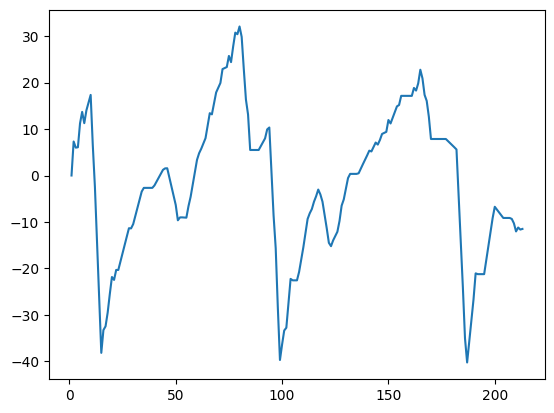}

14

\end{document}